%% file: optimal-beacon-placement.tex
\documentclass[10pt, conference, compsocconf]{IEEEtran}

\input{packages}

\input{notation}

\ifCLASSINFOpdf
\else
\fi

\begin{document}
%
\title{
Towards Optimal Beacon Placement for Range-Aided Localization
}


\author{\IEEEauthorblockN{Ethan Sequeira, Hussein Saad, Stephen Kelly, and Matthew Giamou}
\IEEEauthorblockA{Department of Computing and Software\\
McMaster University\\
Hamilton, Canada\\
\{sequeie, saadh, spkelly, giamoum\}@mcmaster.ca}
}

\maketitle

\begin{abstract}
Range-based localization is ubiquitous: global navigation satellite systems (GNSS) power mobile phone-based navigation, and autonomous mobile robots can use range measurements from a variety of modalities including sonar, radar, and even WiFi signals.  
Many of these localization systems rely on fixed anchors or beacons with known positions acting as transmitters or receivers.
In this work, we answer a fundamental question: given a set of positions we would like to localize, how should beacons be placed so as to minimize localization error? 
Specifically, we present an information-theoretic method for optimally selecting an arrangement consisting of a few beacons from a large set of candidate positions.
By formulating localization as maximum a posteriori (MAP) estimation, we can cast beacon arrangement as a submodular set function maximization problem.
This approach is probabilistically rigorous, simple to implement, and extremely flexible.  
Furthermore, we prove that the submodular structure of our problem formulation ensures that a greedy algorithm for beacon arrangement has suboptimality guarantees.
We compare our method with a number of benchmarks on simulated data and release an open source Python implementation of our algorithm and experiments.

\end{abstract}


%
\IEEEpeerreviewmaketitle

\section{Introduction}
Range-based localization has been successfully deployed at many scales and with a variety of different sensing modalities. 
Prominent successes include the Global Positioning System (GPS)~\cite{hofmann2012global}, and indoor positioning systems based on WiFi signal strength~\cite{raza2021comparing}. 
Accurate localization is essential for safe operation of autonomous mobile robots, and recent work has explored the development of robust and accurate systems that rely on range sensors~\cite{goudar2023continuoustime, papalia2023score, dumbgen2023safe}.
In many industrial and commercial settings, engineers have the opportunity to design an environment which aids state estimation algorithms with fiducial markers or static sensors.
In the case of range-aided localization, budgetary and infrastructural constraints may limit the number or placement of sensing beacons installed in a robot's operational environment.

\begin{figure}
\centering 
\includegraphics[width=\columnwidth]{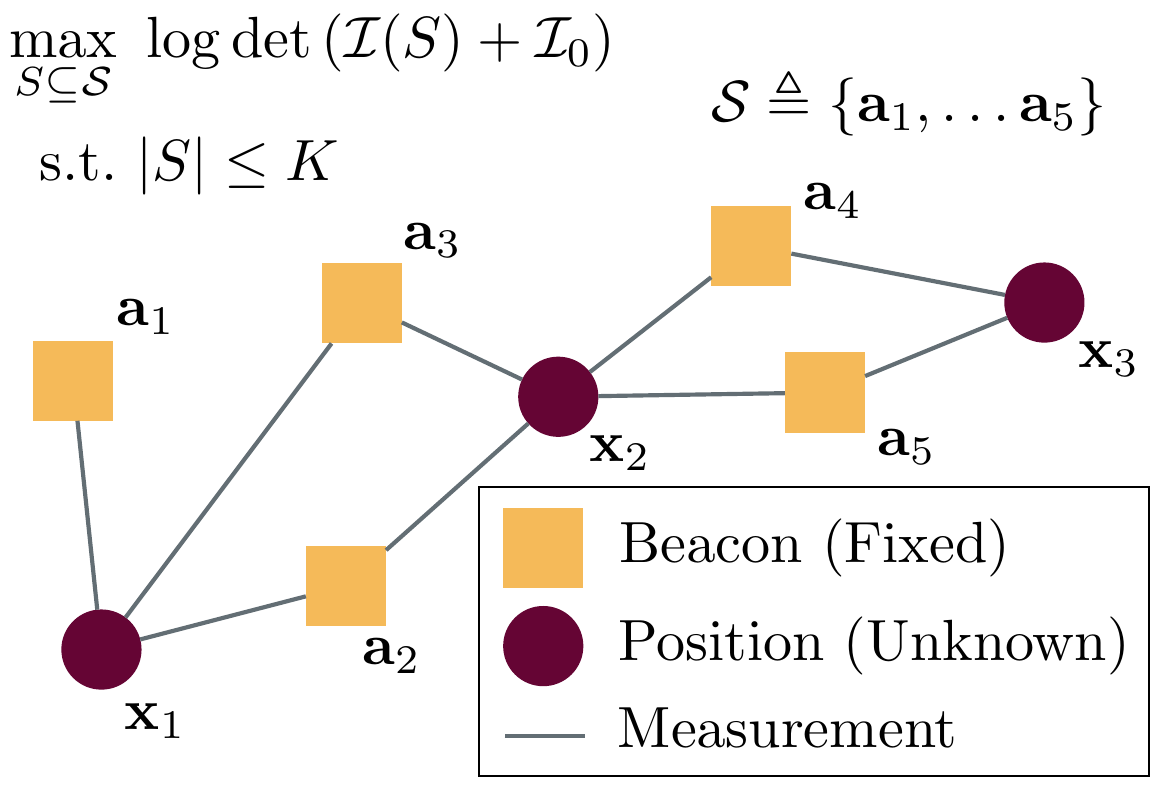}
\caption{We present an information-theoretic approach to arranging the position $\Vector{a}_j$ of $\Budget$ sensing beacons in an environment in which an agent relies on range measurements for localization at points $\Vector{x}_i$ for $i \in \Indices{\NumPositions}$.
	Using a finite set of candidate beacon positions $\BeaconSet$ and an objective function derived with D-optimal Bayesian experiment design, we cast the optimal beacon placement (OBP) problem as submodular set function maximization with suboptimality guarantees.}
\label{fig:main_figure}
\end{figure}

Inspired by the work of Kayhani et al. ~\cite{kayhani2023perceptionaware} on how to optimally place fiducial markers for visual localization in construction environments, we introduce a solution to the \emph{optimal beacon placement} (OBP) problem depicted in \Cref{fig:main_figure}. 
Our problem formulation assumes that the designers of the environment are provided with or able to produce a set of positions at which localization is important.
These may be samples from a trajectory a service robot has to follow, or positions where localization from other sensing modalities is unavailable (e.g., an indoor area without GPS reception).
Furthermore, our formulation incorporates a prior distribution for each position estimate, allowing designers to specify the fact that some positions will have inaccurate onboard estimation (e.g., corridors where visual odometry is highly uncertain in one spatial dimension, or low-friction surfaces that degrade wheel odometry).
Finally, we assume that beacons may only be placed in one of finitely many candidate positions specified by the designer.
In some scenarios, this limitation reflects infrastructural realities, such as the presence or absence of discrete architectural features onto which beacons may be mounted.
In cases without these constraints, our method avoids issues which may arise in a continuous parameterization of OBP's feasible set, while still supporting a fine discretization involving hundreds or thousands of candidate beacon positions.

When combined with a Gaussian model for range measurements, our novel problem formulation suggests a simple greedy algorithm with a provable suboptimality guarantee. 
Our novel contributions are as follows:
\begin{enumerate}
	\item an information-theoretic formulation of OBP for maximum a posteriori (MAP) localization;
	\item a fast greedy solution with a suboptimality guarantee derived from the theory of submodular set function maximization;
	\item simulated experiments comparing the effectiveness of our approach with benchmark algorithms; and 
	\item a publicly available Python repository containing an implementation of our method and experimental results.\footnote{\url{https://github.com/ARCO-Lab/localization-simulator}}
\end{enumerate}
The remainder of this paper is organized as follows: \Cref{sec:related_work} summarizes related work.
In \Cref{sec:range-aided_localization}, we describe our MAP formulation of range-aided localization.
\Cref{sec:beacon_placement} presents OBP, our set function maximization approach to placing a limited number of beacons in an environment.
In \Cref{sec:algorithms}, we demonstrate that a greedy algorithm can find a solution to OBP with a solution within a constant factor of the global optimum, and we test this method against benchmarks in the simulated experiments of \Cref{sec:experiments}.
\Cref{sec:conclusion} summarizes our findings and identifies promising directions for future work.

\section{Related Work} \label{sec:related_work}
State estimation for mobile robots is a rich topic with a long history~\cite{barfoot2017state}. 
In this section, we briefly review recent results in range-based localization and survey prior work related to optimal sensing in robotics and vision.

\subsection{Range-based Localization}
Throughout this paper, we make a distinction between \emph{range-only} localization methods~\cite{kantor2002preliminary}, which don't include information from other sources, and \emph{range-aided} localization, which incorporates other sensing models.
Range-only localization plays an important role in simultaneous localization and mapping (SLAM) for autonomous underwater vehicles~\cite{newman2003pure}, which differ from aerial and terrestrial vehicles in that they cannot rely on optical sensing and are therefore often equipped with acoustic range sensors. 
Recently, convex relaxations~\cite{boyd2004convex} have been applied to the range-based localization problems. 
In \cite{goudar2024optimal} and \cite{papalia2023score}, convex initialization strategies are applied to range-only localization and range-aided SLAM, respectively.
Similarly, convex relaxations methods are used in \cite{dumbgen2023safe} and \cite{papalia2023certifiably} to directly solve these problems and provide certificates of global optimality.
In order to define the optimal beacon placement problem, the present paper uses a formulation of range-aided localization which is compatible with many of the aforementioned solution methods. 
However, the benefits of our approach are agnostic to the actual method used to perform localization. 

Range-only localization has a great deal of overlap with the related problem of sensor network localization (SNL)~\cite{wang2010survey}.
In this work, we use a probabilistic and graph-theoretic problem formulation very similar to the one in \cite{simonetto2014distributed}, which proposes a distributed maximum likelihood solution to SNL.
Additionally, like the SNL literature, we restrict our attention to position estimation, and note that placing multiple range sensors on an agent enables full six degree of freedom pose estimation through range sensing alone~\cite{goudar2023continuoustime}.

\subsection{Optimal Sensor Placement}
The algorithm developed in this work relies on the machinery of submodular set function maximization described in \cite{krause2014submodular}.
This machinery has been successfully applied to a variety of state estimation problems related to robotics including visual feature selection~\cite{carlone2019Attention}, Kalman filtering~\cite{tzoumas2016sensor}, camera placement~\cite{collin_resilient_2019}, and measurement selection for SLAM~\cite{khosoussi2019reliable, tian2018nearoptimala}.
Kayhani et al.~\cite{kayhani2023perceptionaware} use a genetic algorithm to optimally place fiducial tags for visual sensing, and Kaveti et al.~\cite{kaveti2023oasis} consider the inverse problem of optimally arranging cameras on a robot moving through an environment with known static visual features.
The greedy algorithm in \cite{huang2023optimizing} additionally considers the visual information already present in a scene to determine the best locations for fiducial markers. 
While our approach uses an information-theoretic approach and discrete design space similar to these works, we focus exclusively on range-aided localization, and make use of submodularity inherent in our MAP problem formulation to create a greedy algorithm with suboptimality guarantees.

Most prior work on optimal placement of beacons for range sensing focuses on coverage of the environment. 
For two-dimensional localization, Andrews et al.~\cite{andrews2024new} propose a regular lattice which ensures unique localization coverage with a minimal number of beacons. 
Rajagopal et al.~\cite{rajagopal2016beacon} minimize a metric combining localization coverage with geometric dilution of precision (GDOP)~\cite{moragrega2012supermodular}, a quantity proportional to the Cram\'{e}r-Rao lower bound used in this work. 
However, their method does not use an efficient greedy algorithm with suboptimality guarantees.
In \cite{wang2019efficient}, Wang et al. consider a discretized version of the beacon placement problem that is similar to ours. 
They also apply greedy methods, but use a sophisticated submodular set function related to the number of points uniquely localized by a set of beacons. 
Wu et al.~\cite{wu2023optimal} use a quotient-space gradient-based particle swarm optimization (QGPSO) to solve a probabilistic problem formulation similar to ours, but with a focus on accurately characterizing sensor line-of-sight. 
In \cite{zhao2022finding}, emphasis is also placed on modelling line-of-sight, and a local optimization method is used to search their continuous design space.
In contrast to the above works, we focus on a simple information-theoretic objective function with characteristics that can be exploited by greedy algorithms (see \Cref{sec:nms}).
Additionally, none of the prior work discussed in this section considers the specific range-aided MAP localization problem we present in \Cref{sec:MAP}.

Finally, in the realm of SNL, Shames and Summers~\cite{shames2015rigid} study a problem closely related to ours involving the selection of a subset of vertices in a sensor network to specify as anchors with known position.
By using concepts from rigidity theory, they identify a set function maximization formulation of their problem that is \emph{modular}, meaning a greedy algorithm produces a globally optimal solution.
Our problem differs in that we are selecting which anchors to use with a fixed set of vertices that require localization, and we use a purely information-theoretic objective function.
However, analyzing the influence of rigidity on our problem formulation is an intriguing direction for future work. 

\section{Background: Range-Aided Localization} \label{sec:range-aided_localization}
In this section we begin by describing our probabilistic formulation of range-only localization in \Cref{sec:range-only_localization}.
In \Cref{sec:MLE}, we make mild assumptions about range measurement models to describe a maximum likelihood approach to range-only localization. 
In \Cref{sec:MAP}, we incorporate prior estimates of positions to extend the MLE in \Cref{sec:MLE} to a range-aided localization problem. 
By restricting error models to additive Gaussian noise, we reduce range-aided localization to a simple nonlinear least squares problem.

\subsection{Range-Only Localization} \label{sec:range-only_localization}
Consider the problem of estimating an unknown position $\Vector{x} \in \Real^{\Dim}$ using noisy range measurements with $\NumBeacons$ anchors or \emph{beacons} with known positions $\Vector{a}_j \in \Real^{\Dim} \ \forall j \in \Indices{\NumBeacons}$, where $\Indices{n} \Defined \{1, 2, \ldots, n\}$ for any integer $n > 0$.
We restrict our attention to noisy range measurement models of the form 
\begin{equation} \label{eq:measurement_model}
	g_j(\Vector{x}) = \|\Vector{x} - \Vector{a}_j\|_2 + \eta_j,
\end{equation}
where $\eta_j \sim \NormalDistribution{\Zero}{\sigma_j^2}$ is additive zero-mean Gaussian noise with variance $\sigma_j^2$ that may be a function of the particular beacon $j$ and its position $\Vector{a}_j$ with respect to $\Vector{x}$.\footnote{As an example, the noise may depend on the range $\|\Vector{x} - \Vector{a}_j\|_2$.}
Additionally, we may have a prior estimate encoded as a multivariate Gaussian distribution with mean $\Prior{\Vector{x}} \in \Real^{\Dim}$ and covariance $\Prior{\Covariance} \in \PD{\Dim}$,\footnote{We denote prior estimates with $\Prior{\cdot}$ and posterior estimates gleaned from measurements with $\Estimate{\cdot}$.} where $\PD{n}$ ($\PSD{n}$) is the space of $n\times n$ symmetric positive (semi)definite matrices.

\begin{figure}
\centering 
\includegraphics[width=\columnwidth]{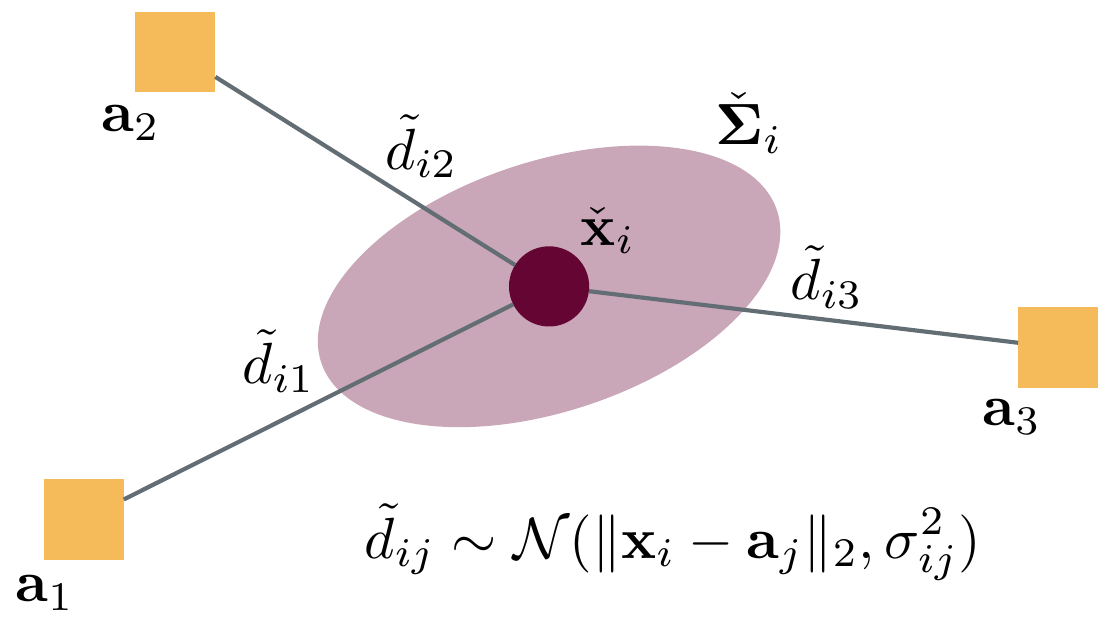}
\caption{Range-aided localization of a single unknown position $\Vector{x}_i$ with prior mean $\Prior{\Vector{x}}_i$ and covariance $\Prior{\Matrix{\Sigma}}_i$. For each beacon $\Vector{a}_j$ in range, a noisy measurement $\tilde{d}_{ij}$ is received.}
\label{fig:localization}
\end{figure}

\subsection{Maximum Likelihood Estimation} \label{sec:MLE}
In this section, we use the probabilistic measurement model in \Cref{eq:measurement_model} to develop a maximum likelihood estimate (MLE) $\Estimate{\Vector{x}}$ of $\Vector{x}$~\cite{barfoot2017state}.
We begin by noting that the conditional probability of observing a particular measurement is 
\begin{equation}
	\tilde{d}_j \sim p_j(\cdot|\Vector{x}) = \NormalDistribution{\|\Vector{x} - \Vector{a}_j\|_2}{\sigma_j^2}.
\end{equation}
Since the measurements are independent, their joint conditional distribution is the product
\begin{equation}
	p(\tilde{D}|\Vector{x}) \Define \prod_{j=1}^{\NumBeacons} p_j(\tilde{d}_j|\Vector{x}),
\end{equation}
where $\tilde{D} \Define \{\tilde{d}_j\}$ is the set of all measurements.
The MLE estimator of $\Vector{x}$ is therefore 
\begin{equation}
\Estimate{\Vector{x}} \Defined \ArgMax{\Vector{x}\in \Real^\Dim} \ p(\tilde{D}|\Vector{x}).
\end{equation}
Taking the negative $\log$ of the likelihood, we can obtain $\Estimate{\Vector{x}}$ by solving the following equivalent nonlinear least squares minimization problem:
\begin{problem}[MLE for Range-Only Localization] \label{prob:mle}
For noisy range measurements $\tilde{d}_j \in \tilde{D}$ with associated variance $\sigma_j^2$, find $\Estimate{\Vector{x}}$ that solves
\begin{equation}
	\min_{\Vector{x}\in \Real^\Dim} \sum_{j=1}^{\NumBeacons} \frac{1}{\sigma_j^2}(\|\Vector{x} - \Vector{a}_j\|_2 - \tilde{d}_j)^2,
\end{equation}
where each $\Vector{a}_j$ is a sensing beacon with a known fixed position. 
\end{problem}

The Cram\'{e}r-Rao lower bound (CRLB) provides a lower bound on the covariance of the MLE estimator $\Estimate{\Vector{x}}$~\cite{barfoot2017state}:
\begin{equation} \label{eq:cramer-rao}
	\mathrm{Cov} (\Estimate{\Vector{x}}) \succeq \FIM(\Vector{x})^{-1}, 
\end{equation}
where we have used the \emph{Fisher information matrix} (FIM):
\begin{equation} \label{eq:fisher}
	\FIM(\Vector{x}) \Defined \Expectation{\nabla_{\Vector{x}} \log p(\tilde{D}|\Vector{x})(\nabla_{\Vector{x}} \log p(\tilde{D}|\Vector{x}))^\top}{\tilde{D}}.
\end{equation}
In \Cref{sec:information_theoretic}, we use the FIM to define an information-theoretic objective function of beacon positions $\Vector{a}_j$.

\subsection{Maximum a Posteriori Estimation} \label{sec:MAP}
Suppose that we also have access to a prior distribution $p(\Vector{x}) \sim \NormalDistribution{\Prior{\Vector{x}}}{\Prior{\Covariance}}$ for our unknown position $\Vector{x}$.
This may come from some other independent noisy localization system (e.g., wheel odometry), or can function as a way of specifying that an agent is expected to be in a certain area at a certain time as part of its operation. 
Using Bayes' rule, we obtain the following MAP estimation problem~\cite{barfoot2017state}:
\begin{problem}[Range-Aided MAP Localization] \label{prob:map}
Range-aided MAP localization with a Gaussian prior takes the form
\begin{equation}
	\min_{\Vector{x}\in \Real^\Dim} \|\Vector{x} - \Prior{\Vector{x}}\|^2_{\Prior{\Covariance}} + \sum_{j=1}^{\NumBeacons} \frac{1}{\sigma_j^2}(\|\Vector{x} - \Vector{a}_j\|_2 - \tilde{d}_j)^2,
\end{equation}
where $\|\cdot \|_{\Matrix{A}}$ denotes the Mahalanobis distance with respect to $\Matrix{A} \in \PSD{n}$~\cite{barfoot2017state}.
\end{problem}

\Cref{prob:map} is essentially \Cref{prob:mle} with a regularizing term defined by the prior.
This regularizing term is important to the method developed in \Cref{sec:beacon_placement}.
Both the MLE and MAP problems are smooth, unconstrained, and nonconvex optimization problems which can be efficiently and reliably searched for local minima by Newton's method and its many variants. 

\section{Optimal Beacon Placement} \label{sec:beacon_placement}
In this section, we use the probabilistic range-aided localization problem of \Cref{sec:range-aided_localization} to introduce the optimal beacon placement problem.
\Cref{sec:feasible} describes two ways of parameterizing this problem's design space, and in \Cref{sec:information_theoretic} we define an information-theoretic notion of optimality. 
Finally, \Cref{sec:OBP} introduces OBP as a set function maximization problem which we will solve in \Cref{sec:algorithms}.

\subsection{Feasible Beacon Positions} \label{sec:feasible}
Suppose that we are designing an environment for an agent equipped with a sensor capable of measuring its range to fixed beacons with known position.  
Furthermore, we are constrained to a budget of $\Budget \geq 1$ beacons.
Physical constraints may limit the placement of beacons to some subset $\FreeSpace \subset \Real^\Dim$. 
We can therefore completely specify all feasible arrangements of beacons via $\Vector{a}_j \in \FreeSpace \ \forall j \in \Indices{\Budget}$. 

Alternatively, we may be constrained to mount beacons on some finite $\BeaconSet \subset \FreeSpace$, where $\Cardinality{\BeaconSet} = \NumBeacons \gg \Budget$.
In this \emph{discrete} formulation, there are $\binom{\NumBeacons}{\Budget} \Defined \frac{\NumBeacons!}{\Budget!(\NumBeacons - \Budget)!}$ candidate beacon arrangements. 
This scenario may reflect real infrastructural limitations on the placement of beacons, but we will see in \Cref{sec:algorithms} that this combinatoric view of the problem also serves as a practical algorithmic contrivance.

In our beacon arrangement problem, we aim to estimate $\NumPositions$ unknown positions $\Vector{x}_i \ \forall i \in \Indices{\NumPositions}$, each with a prior Gaussian distribution $p_i(\Vector{x}_i) \sim \NormalDistribution{\Prior{\Vector{x}}_i}{\Prior{\Covariance}_i}$
Due to range limits or occlusions, only a subset of the beacons $a_j$ can be used at each position $i$ (see \Cref{fig:main_figure}).
We encode this information in a bipartite graph $\Graph = (\Vertices, \Edges)$, where $\Vertices \Defined \Indices{\NumPositions} \cup \Indices{\NumBeacons}$ and each edge $i, j \in \Edges$ is identified with a measurement $\tilde{d}_{ij}$.
The neighbourhood function which maps position $i$ to the beacons it can measure is denoted
\begin{equation}
	\Neighbourhood: \Indices{\NumPositions} \rightarrow 2^{\Indices{\NumBeacons}}.
\end{equation} 
Just as in \Cref{eq:measurement_model}, each position $i$ has an associated set of measurement models
\begin{equation} \label{eq:measurement_model_full}
	g_{ij}(\Vector{x}_i) = \|\Vector{x}_i - \Vector{a}_j\|_2 + \eta_{ij} \ \forall j \in \Neighbourhood(i), 
\end{equation}
and we denote the distribution of measurement $\tilde{d}_{ij}$ with $p_{ij}(\cdot | \Vector{x}_i)$. 
The estimation of each unknown position $\Vector{x}_i$ is independent of the other positions, but the placement of each beacon $j$ affects the localization quality for multiple positions. 

For each $S \subseteq \BeaconSet$, consider the binary vector $s \in \{0, 1\}^{\NumBeacons}$ indicating membership in $S$:
\begin{equation}
	s_j = \begin{cases}
		1, \ \Vector{a}_j \in S, \\
		0, \ \mathrm{ otherwise.}
	\end{cases}
\end{equation}
This vector allows us to conveniently write our MAP likelihood as a function of included beacons:
\begin{equation} \label{eq:map_conditioned}
	p(\tilde{D} | X; s) \Defined p(\Vector{x}) \prod_{i=1}^{\NumPositions} \prod_{j\in \Neighbourhood(i)} p_{ij}(\tilde{d}_{ij}| \Vector{x}_i)^{s_j},
\end{equation}
where we abuse our notation and use $\Vector{x} \in \Real^{\Dim \NumPositions}$ to refer to the concatenation of all position $\Vector{x}_i$ in \Cref{eq:map_conditioned} onwards.
We will also make the simplifying assumption that the prior belief $p(\Vector{x}_i)$ for position $\Vector{x}_i$ is independent of the others, i.e., 
\begin{equation} \label{eq:prior_independence}
	p(\Vector{x}) = \prod_{i=1}^{\NumPositions} p(\Vector{x}_i).
\end{equation}
This assumption may not be true for prior estimates originating from visual SLAM or a similar method, but can still serve as a useful approximation. 
Furthermore, this assumption of independence for each position's prior enables a very efficient solution method to the optimal beacon placement problem in \Cref{sec:algorithms}.

\subsection{An Information-Theoretic Objective} \label{sec:information_theoretic}
In order to define an \emph{optimal} placement of beacons over $\FreeSpace$ or $\BeaconSet$, we need to introduce a sensible objective function.
Recall the FIM of \Cref{eq:fisher}, whose inverse is a lower bound to the MLE $\Estimate{\Vector{x}}$. 
This motivates us to choose an objective function that \emph{maximizes} the eigenvalues of $\FIM(\Vector{x})$.
Since $\FIM(\Vector{x}) \in \PSD{\Dim \Budget}$, we can select the \emph{D-criterion} as our objective function~\cite{carlone2019Attention}:
\begin{equation} \label{eq:log_det}
	\CostFunction(S) \Defined \log \det \left(\FIM(S) + \FIM_0\right),
\end{equation}
where 
\begin{equation} \label{eq:FIM_init}
	\FIM_0 \Defined \mathrm{Diag}( \{\Prior{\Covariance}_i^{-1}\}_{i=1}^\NumPositions),
\end{equation}
and $\mathrm{Diag}$ constructs a block-diagonal matrix with its input.
In \Cref{eq:log_det}, $\FIM$ has been overloaded to take $S \subseteq \BeaconSet$ as its argument:
\begin{equation} \label{eq:fisher_set_function}
	\FIM(S) \Defined \sum_{s\in S} \Expectation{\nabla_{\Vector{x}} \log p(\tilde{D}|\Vector{x}; s)(\nabla_{\Vector{x}} \log p(\tilde{D}|\Vector{x}; s))^\top}{\tilde{D}}, 
\end{equation}
and 
\begin{equation} \label{eq:log_likelihood}
\begin{aligned}
	\log p(\tilde{D}|\Vector{x}; s) &= \sum_{i=1}^{\NumPositions} \sum_{j\in \Neighbourhood(i)} s_j \log p(\tilde{d}_{ij}|\Vector{x}_i) \\
	&= \sum_{i=1}^{\NumPositions} \sum_{j\in \Neighbourhood(i)} \frac{s_j}{\sigma_{ij}^2} (\|\Vector{x}_i - \Vector{a}_j\|_2 - \tilde{d}_j)^2. 
\end{aligned}
\end{equation}
Without the $\NumPositions$ terms corresponding to the information in the prior distribution, maximizing $\CostFunction$ is akin to minimizing the lower bound (via the CRLB) of the \emph{volume} of a fixed level set of the MLE solution to \Cref{prob:mle}'s distribution. 
By incorporating the prior, we instead minimize uncertainty of the posterior normal approximation in the MAP formulation~\cite[Sec. 4]{chaloner1995bayesian}.

When actually computing \Cref{eq:fisher_set_function}, we will approximate the expectation with the one-sample mean consisting of $\nabla_{\Vector{x}} \log p(\tilde{D}|\Vector{x}; s)$ evaluated at the observed measurements $\tilde{D}$~\cite{kaveti2023oasis}. 
Since we are proposing an offline design procedure in simulation, we will also have access to the  ground truth value of the parameter $\bar{\Vector{x}}$. 
Finally, since the measurements and prior distribution for each position with index $i \in \Indices{\NumPositions}$ are independent of all other positions, the expression for the FIM in \Cref{eq:fisher_set_function} has a block diagonal structure, and the evaluation of \Cref{eq:log_det} can be efficiently implemented as the sum of the log-determinant of each diagonal submatrix.

\subsection{D-Optimal Beacon Placement} \label{sec:OBP}
We are now ready to introduce our problem of interest: 

\begin{problem}[Optimal Beacon Placement (OBP)] \label{prob:optimal_beacon_placement}
For a finite set of candidate beacon positions $\BeaconSet \subset \Real^\Dim$, solve 
\begin{equation}
\begin{aligned}
	\max_{S \subseteq \BeaconSet} \ & \CostFunction(S) \\
	\mathrm{s.t.} \ & \Cardinality{S} \leq K,
\end{aligned}
\end{equation}
where $\CostFunction$ is the information-theoretic cost function defined in \Cref{eq:log_det}.
\end{problem}
\noindent \Cref{prob:optimal_beacon_placement} is a \emph{set function maximization} problem and is NP-hard in general~\cite{krause2014submodular}.

\section{Algorithms} \label{sec:algorithms}
In this section, we use the properties of $\CostFunction$ to derive a greedy algorithm with suboptimality guarantees. 

\subsection{Normalized, Monotone, and Submodular Set Functions} \label{sec:nms}
We begin with a few simple definitions for nonnegative set functions $f: 2^{\BeaconSet} \rightarrow \Real_+$ over a finite set $\BeaconSet$. 
A set function is \emph{normalized} if $f(\emptyset) = 0$, where $\emptyset = \{\}$ is the empty set. 
A set function is \emph{monotone} if $f(A) \leq f(B)$ for any $A \subseteq B$.
Finally, a set function is \emph{submodular} if 
\begin{equation}
	f(A \cup \{e\}) - f(A) \geq f(B \cup \{e\}) - f(B)
\end{equation}
for any $A \subseteq B \subset \BeaconSet$ and $e \in \BeaconSet \backslash B$~\cite{krause2014submodular}.
When $f$ is normalized, monotone, and submodular (NMS), the following helpful result holds. 

\begin{theorem}[Greedy Maximization of NMS Functions~\cite{krause2014submodular}] \label{thm:submodularity}
Let $S_g \subset \BeaconSet$ be the greedy solution given by \Cref{alg:greedy} to a set maximization problem like \Cref{prob:optimal_beacon_placement} for some $K = \Cardinality{S_g}$ and a set function $f$.
If $f$ is NMS, then 
\begin{equation}
	f(S_g) \geq (1 - 1/e)f(S^\star) \approx 0.63 f(S^\star), 
\end{equation}
where $S^\star$ is the optimal solution. 
\end{theorem}

\Cref{thm:submodularity} gives a suboptimality bound for the greedy solution of NMS set function maximization problems. 
Recall that a brute-force solution to \Cref{prob:optimal_beacon_placement} requires the evaluation of $\binom{\NumBeacons}{\Budget}$ subsets, whereas the greedy \Cref{alg:greedy} has a linear time complexity of $O(\Budget\NumBeacons)$. 

\begin{small}
  \begin{algorithm}[t]
    \caption{Greedy Set Function Maximization \label{alg:greedy}}
    \begin{algorithmic}[1]
      \Input A set function $f$, the set $\BeaconSet$, a budget $\Budget < \Cardinality{\BeaconSet}$
      \Output An approximate solution to \Cref{prob:optimal_beacon_placement}
      \Function{Greedy}{}
      \State $S \leftarrow \emptyset$
      \Comment{Initialize the solution set}
      \While{$\Cardinality{S} \leq \Budget$} \Comment{Greedily select $\Budget$ elements}
      	\State $e^\star = \ArgMax{e \in \BeaconSet \backslash S} f(S \cup \{e\})$
      	\State $S \leftarrow S \cup \{e^\star\}$
      \EndWhile
      \State \Return $S$
      \EndFunction
    \end{algorithmic}
  \end{algorithm}
\end{small}

\subsection{Analyzing $\CostFunction$}
In this section, we demonstrate that a constant modification of $\CostFunction$ is in fact NMS and therefore benefits from \Cref{thm:submodularity}. 
First note that 
\begin{equation}
\begin{aligned}
	\CostFunction(\emptyset) &= \log \det \FIM_0 \\
	&= \log \det \left(\mathrm{Diag}( \{\Prior{\Covariance}_i^{-1}\}_{i=1}^\NumPositions)\right) \\
	& = \log \det\left(\sum_{i=1}^\NumPositions \Prior{\Covariance}_i^{-1}\right),
\end{aligned}
\end{equation}
which is well-defined so long as $\sum_{i=1}^\NumPositions \Prior{\Covariance}_i^{-1} \succ 0$.
Therefore, the biased objective function
\begin{equation}
	\CostFunctionNMS(S) \Defined \CostFunction(S) - \CostFunction(\emptyset)
\end{equation}
is normalized and has the same maximizer as $\CostFunction$ with respect to \Cref{prob:optimal_beacon_placement}.

Next, we observe that our approximate computation of the FIM involves adding a set of rank-one updates for each selected $e \in S \subset \BeaconSet$. 
Shamaiah et al.~\cite{shamaiah2010greedy} demonstrated that the $\log \det$ of sums of this form are in fact monotone submodular functions. 
Therefore, we conclude that $\CostFunctionNMS$ is NMS and can be greedily maximized for a solution that satisfies the suboptimality bound in \Cref{thm:submodularity}.

\subsection{CMA-ES Adaptation}
Inspired by the use of a genetic algorithm in \cite{kayhani2023perceptionaware} for fiducial tag placement, in this section we consider an alternative to greedy set function maximization from the evolutionary optimization literature.
Specifically, we apply the covariance matrix adaptation evolution strategy (CMA-ES) to OBP. 
CMA-ES is a robust evolutionary algorithm for nonlinear, nonconvex optimization problems~\cite{hansen2016cmaes}. 
Given an objective function, CMA-ES searches for an optimal solution by iteratively updating a multivariate normal distribution $\cal{N}$$(\Vector{a}, \Covariance_{a})$ of candidate solutions $\Vector{a} \in \Real^{\Dim \Budget}$, based on the covariance matrix $\Covariance_{a}$ of the best solutions found in previous iterations. 
For the OBP, the objective function is represented by the information gain function $\CostFunction$ in \Cref{eq:log_det}. 
The algorithm halts primarily based on convergence criteria that determine whether further iterations are likely to lead to significant improvements in the objective function. 
Since CMA-ES operates on continuous domains, we adapt its solutions to our discrete problem space by selecting the $K$ beacons in $\BeaconSet$ closest (in a Euclidean sense) to the positions returned in its output $\Vector{a}$. 


\section{Experiments} \label{sec:experiments}
In this section, we systematically evaluate the performance of various algorithms for solving the OBP. 
All experiments were performed on a machine with an Intel(R) Xeon(R) W-11955M CPU @ 2.60 GHz, and 64.0 GB of RAM.
We begin by describing benchmark algorithms in \Cref{sec:benchmarks}.
\Cref{sec:brute_force} contains a small empirical demonstration of our greedy algorithm's guaranteed suboptimality bounds.
We conclude our experiments with larger-scale comparisons of our greedy method, CMA-ES, and the benchmark algorithms in \Cref{sec:large_scale}, presenting statistics for runtime and the localization accuracy resulting from beacons selected by each method. 

The value of $\Prior{\sigma}_i^2$ varies between experiments, but within each experiment it is the same for all $i \in \Indices{\NumPositions}$. 
Finally, for each experimental trial, the ground truth positions $\bar{\Vector{x}}_i$ are randomly sampled from their prior distribution $p_i(\Vector{x}_i)$. 

\begin{figure}[h!]
    \centering
    \includegraphics[width=\columnwidth]{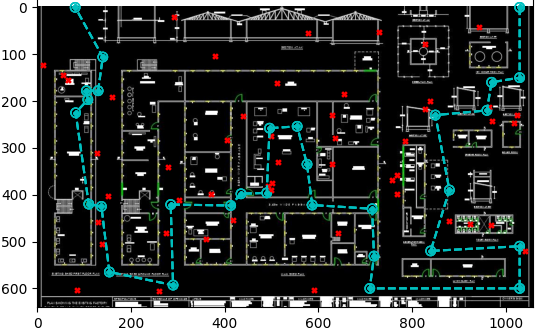}
    \caption{Blueprints of a factory setting retrieved from \cite{factoryWebsite} overlaid with a simulated robot trajectory in cyan and candidate beacon positions in red. Environments using this trajectory and random samplings of candidate beacon positions are used in the experiments of \Cref{sec:large_scale}.}
    \label{fig:factory}
\end{figure}

\subsection{Benchmarks} \label{sec:benchmarks}
In addition to the greedy information-theoretic method (henceforth called \textit{Greedy}) and CMA-ES algorithm introduced in \Cref{sec:algorithms}, we include the following benchmark algorithms in our experiments:
\begin{enumerate}
	\item \textit{Random}: Randomly selects a subset of $\Budget$ beacons.
	\item \textit{Brute-Force}: Exhaustively searches all possible subsets of $\Budget$ beacons and returns the subset with highest information gain.
	\item \textit{Measurement Greedy}: Greedily selects beacons so as to maximize the total number of distance measurements received across all positions $\Vector{x}_i$. 
	\item \textit{Coverage Greedy}: Begins by greedily selecting beacon positions to ensure coverage of all positions $\Vector{x}_i$. Once all positions are covered, it follows an approach identical to \textit{Measurement Greedy}, greedily selecting beacons which maximize the number of distance measurements. This algorithm ensures (if possible) that all positions receive at least one measurement before optimizing for total measurements received by all positions.
\end{enumerate}

Once a subset of beacons is selected, \Cref{prob:map} is solved with Newton's method~\cite{nocedal2006numerical} to yield $\Estimate{\Vector{x}}$.
Since we are not concerned with evaluating solution methods for \Cref{prob:map} and want to avoid local minima in its nonconvex objective function, we use the ground truth value of $\bar{\Vector{x}}$ to initialize Newton's method.
The root mean squared error (RMSE) is used to compare and evaluate the accuracy of the MLE solution:
\begin{equation}
	\mathrm{RMSE}(\Estimate{\Vector{x}}) \Define \sqrt{\frac{1}{\NumPositions}\sum_{i=1}^\NumPositions \|\Estimate{\Vector{x}}_i - \bar{\Vector{x}}_i\|^2}.
\end{equation}


\subsection{Comparing Brute-Force and Greedy} \label{sec:brute_force}
\begin{figure}[!h]
    \centering
    \includegraphics[width=\columnwidth]{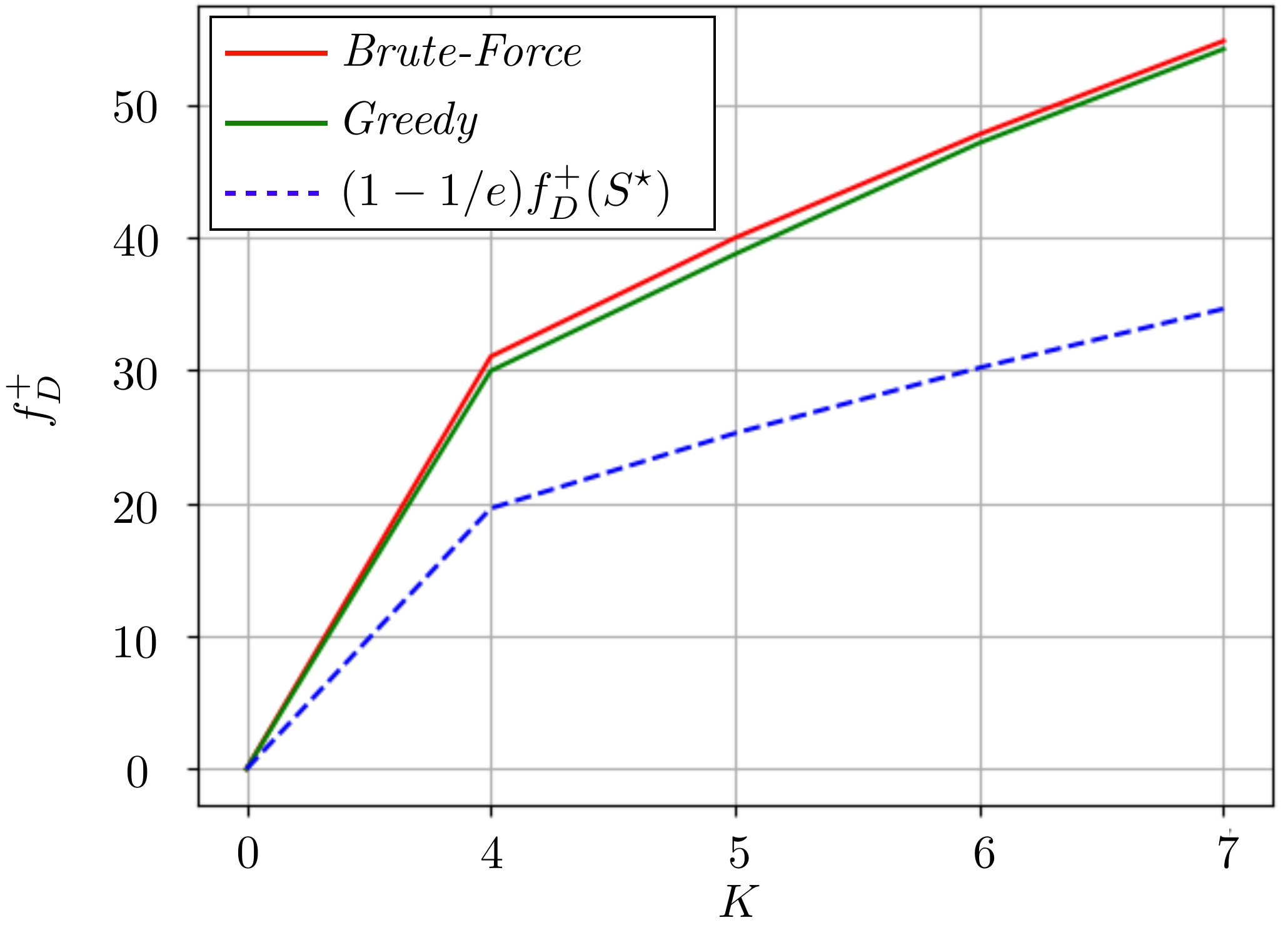}
    \caption{Information-theoretic objective function $\CostFunctionNMS$ with respect to the number of selected beacons $\Budget$. The dashed blue line is the suboptimality bound of \Cref{thm:submodularity}. For all values of $\Budget$ tested, our \textit{Greedy} method satisfies the suboptimality bound and is very close to the globally optimal solution computed by \textit{Brute-Force}.}
    \label{fig:increasingK}
\end{figure}

Our first experiment compares the performance of the \textit{Brute-Force} and \textit{Greedy} algorithms for $\Budget = 1,\ldots, 7$ in a scenario with $\NumBeacons=20$ candidate beacons and $\NumPositions=10$ positions in $\Dim=3$ dimensions.
Because of the combinatoric runtime complexity of the \textit{Brute-Force} method, we were limited to small values of $\Budget$ and $\NumBeacons$. 
\Cref{fig:increasingK} empirically verifies that the suboptimality guarantees of \Cref{thm:submodularity} do in fact apply to our information theoretic cost function $\CostFunctionNMS$. 
The results in \Cref{fig:increasingK} are for a cherrypicked set of candidate beacon positions which demonstrated the potential for a gap in performance between the \textit{Greedy} and \textit{Brute-Force} methods: the vast majority of randomly sampled beacon positions result in the \textit{Greedy} algorithm recovering the globally optimal solution (in a fraction of the time required by \textit{Brute-Force}). 

\subsection{Large Scale Experiment} \label{sec:large_scale}
In this section, we compare our algorithms for solving OBP on a larger 2D problem instance.
\Cref{fig:factory} illustrates a plausible trajectory for a robot within a factory. 
The red cross markers are $\NumBeacons=50$ candidate beacon positions $\Vector{a}_j \in \BeaconSet$, the blue dots are $\NumPositions = 30$ robot positions $\Vector{x}_i$ forming a trajectory, and the blue circles around positions are equiprobable contours corresponding to one standard deviation for their isotropic prior distributions with variance $\Prior{\Covariance}_i = \Prior{\sigma}_i^2 \Identity$.
We use a constant range measurement variance of $\sigma_{ij}^2 = 25$ m$^2$ across all trials, and we introduce a ``cutoff" radius parameter $C > 0$ which adds sparsity to the bipartite measurement graph $\Graph$ by preventing range measurements between positions $\Vector{x}_i$ and beacons $\Vector{a}_j$ separated by a Euclidean distance greater than $C$. 

For each set of the parameter values $\Budget$, $C$, and $\Prior{\sigma}_i$, 50 random trials (i.e., realizations of random ground truth samples $\bar{\Vector{x}}$, random beacon positions $\Vector{a}_j$, and noisy measurements $\tilde{D}$) in the environment displayed in \Cref{fig:factory} were generated, and each of the four algorithms described in \Cref{sec:benchmarks} was used to solve the resulting instance of \Cref{prob:optimal_beacon_placement}. 

\begin{figure*}[htb]
    \centering
    \includegraphics[width=\textwidth]{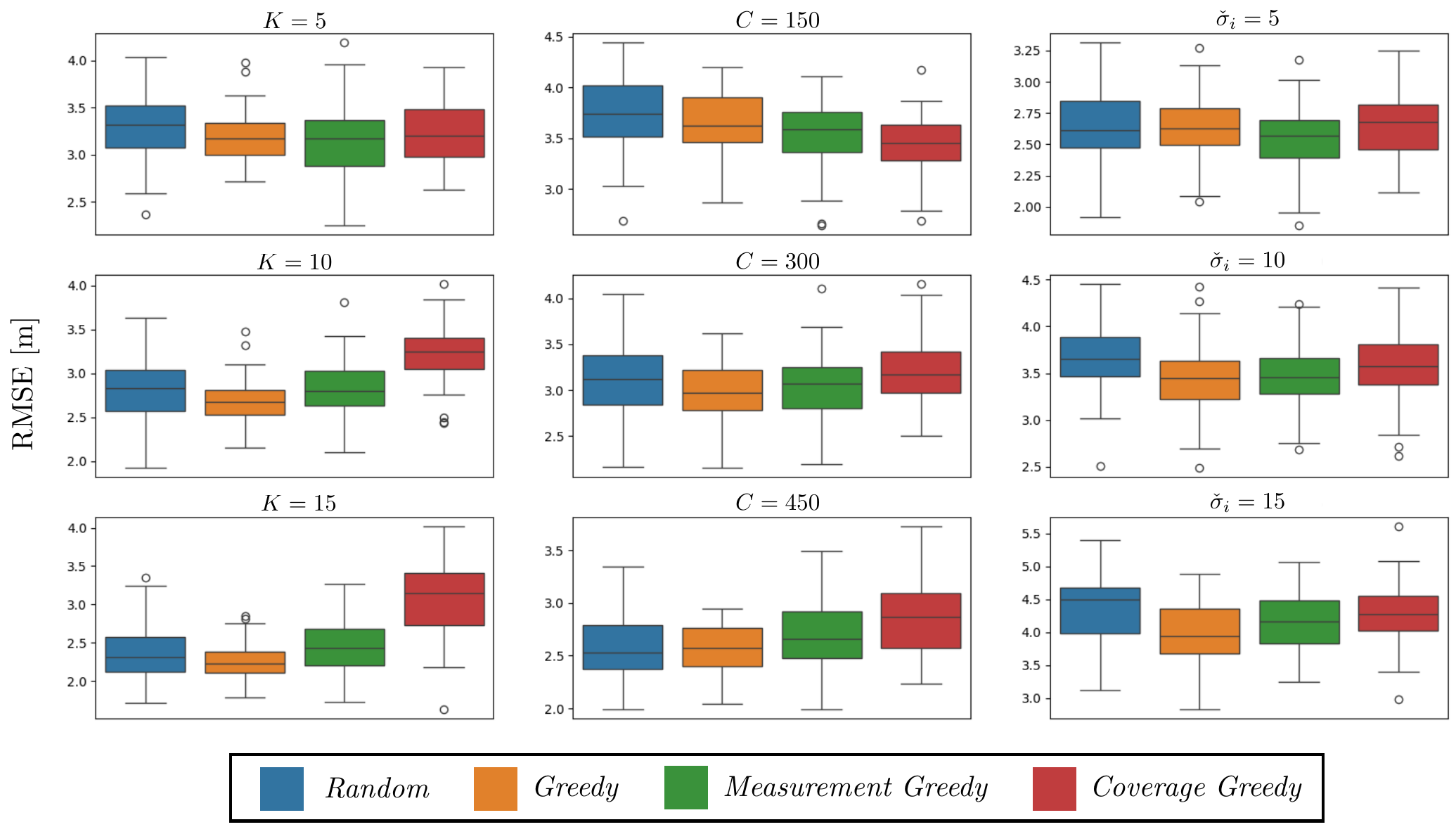}
    \caption{Boxplots of RMSE in meters for experiment in \Cref{sec:large_scale}. Each subplot summarizes the distribution of the RMSE over 50 random trials. Where unspecified, parameters are set to their default values of $K=5$, $C=250$, and $\Prior{\sigma}_i=8$.}
    \label{fig:rmse_results}
    \vspace{-3mm}
\end{figure*}

\Cref{fig:rmse_results} summarizes the distribution of the RMSE for each algorithm for a variety of parameter values.
Where unspecified, baseline values of $\Budget=5$, $C=250$, and $\Prior{\sigma}_i=8$ were used. 
The RMSE and runtime of the same set of experiments are displayed in \Cref{tab:algorithm_performance}, with parameters changed from their baseline values indicated in the leftmost column.

\begin{table}[h]
\centering
\caption{Experimental results for \Cref{sec:large_scale}.}
\label{tab:algorithm_performance}
\scalebox{0.65}{
\begin{tabular}{llllllll}

 & \multicolumn{2}{c}{Algorithm} & \multicolumn{2}{c}{RMSE} & \multicolumn{2}{c}{Runtime} \\
& & & mean [m] & std [m] & mean [s] & std [s] \\
\midrule
\multirow{4}{*}{$K=5$} & \textit{Random} &  & 3.30 & 0.374 & 4.35 $\times 10^{-3}$ & 9.57 $\times 10^{-4}$ \\
& \textit{Greedy} &  & 3.19 & 0.275 & 0.527 & 0.0670 \\
& \textit{Measurement Greedy} &  & \textbf{3.14} & 0.384 & 4.87 $\times 10^{-3}$ & 5.25 $\times 10^{-4}$ \\
& \textit{Coverage Greedy} &  & 3.21 & 0.336 & 6.08 $\times 10^{-3}$ & 1.34 $\times 10^{-3}$ \\
\midrule
\multirow{4}{*}{$K=10$} & \textit{Random} &  & 2.82 & 0.359 & 0.0168 & 4.32 $\times 10^{-3}$ \\
& \textit{Greedy} &  & \textbf{2.67} & 0.257 & 3.27 & 0.432 \\
& \textit{Measurement Greedy} &  & 2.82 & 0.324 & 0.0191 & 4.98 $\times 10^{-3}$ \\
& \textit{Coverage Greedy} &  & 3.23 & 0.350 & 9.91 $\times 10^{-3}$ & 3.48 $\times 10^{-3}$\\
\midrule
\multirow{4}{*}{$K=15$} & \textit{Random} &  & 2.37 & 0.358 & 0.0330 & 2.75 $\times 10^{-3}$ \\
& \textit{Greedy} &  & \textbf{2.25} & 0.250 & 10.0 & 0.899 \\
& \textit{Measurement Greedy} &  & 2.44 & 0.366 & 0.0368 & 3.88 $\times 10^{-3}$\\
& \textit{Coverage Greedy} &  & 3.07 & 0.518 & 0.0141 & 0.0107 \\
\midrule
\multirow{4}{*}{$C=150$} & \textit{Random} &  & 3.73 & 0.370 & 4.03 $\times 10^{-3}$ & 1.09 $\times 10^{-3}$\\
& \textit{Greedy} &  & 3.63 & 0.319 & 0.501 & 0.123 \\
& \textit{Measurement Greedy} &  & \textbf{3.52} & 0.348 & 4.48 $\times 10^{-3}$ & 1.08 $\times 10^{-3}$ \\
& \textit{Coverage Greedy} &  & 3.44 & 0.290 & 6.36 $\times 10^{-3}$& 1.57 $\times 10^{-3}$\\
\midrule
\multirow{4}{*}{$C=300$} & \textit{Random} &  & 3.11 & 0.418 & 5.05 $\times 10^{-3}$ & 1.25 $\times 10^{-3}$\\
& \textit{Greedy} &  & \textbf{3.00} & 0.318 & 0.612 & 0.132 \\
& \textit{Measurement Greedy} &  & 3.04 & 0.373 & 5.62 $\times 10^{-3}$ & 1.24 $\times 10^{-3}$ \\
& \textit{Coverage Greedy} &  & 3.21 & 0.362 & 5.68 $\times 10^{-3}$& 1.61 $\times 10^{-3}$ \\
\midrule
\multirow{4}{*}{$C=450$} & \textit{Random} &  & 2.59 & 0.305 & 5.58 $\times 10^{-3}$ & 9.56 $\times 10^{-4}$ \\
& \textit{Greedy} &  & \textbf{2.55} & 0.231 & 0.689 & 0.134 \\
& \textit{Measurement Greedy} &  & 2.70 & 0.319 & 6.28 $\times 10^{-3}$ & 1.12 $\times 10^{-3}$ \\
& \textit{Coverage Greedy} &  & 2.88 & 0.361 & 5.10 $\times 10^{-3}$& 2.11 $\times 10^{-3}$ \\
\midrule
\multirow{4}{*}{$\Prior{\sigma_i}=5$} & \textit{Random} &  & 2.64 & 0.300 & 4.74 $\times 10^{-3}$ & 1.29 $\times 10^{-3}$ \\
& \textit{Greedy} &  & 2.63 & 0.242 & 0.557 & 0.129 \\
& \textit{Measurement Greedy} &  & \textbf{2.54} & 0.269 & 5.26 $\times 10^{-3}$& 1.27 $\times 10^{-3}$ \\
& \textit{Coverage Greedy} &  & 2.64 & 0.251 & 6.59 $\times 10^{-3}$ & 1.97 $\times 10^{-3}$ \\
\midrule
\multirow{4}{*}{$\Prior{\sigma_i}=10$} & \textit{Random} &  & 3.68 & 0.396 & 4.32 $\times 10^{-3}$ & 6.73 $\times 10^{-4}$ \\
& \textit{Greedy} &  & \textbf{3.45} & 0.360 & 0.525 & 0.0449 \\
& \textit{Measurement Greedy} &  & \textbf{3.45} & 0.336 & 5.14 $\times 10^{-3}$ & 1.09 $\times 10^{-3}$ \\
& \textit{Coverage Greedy} &  & 3.54 & 0.375 & 6.23 $\times 10^{-3}$& 1.59 $\times 10^{-3}$\\
\midrule
\multirow{4}{*}{$\Prior{\sigma_i}=15$} & \textit{Random} &  & 4.36 & 0.523 & 4.23 $\times 10^{-3}$ & 5.73 $\times 10^{-4}$\\
& \textit{Greedy} &  & \textbf{3.98} & 0.411 & 0.527 & 0.0840 \\
& \textit{Measurement Greedy} &  & 4.13 & 0.396 & 4.93 $\times 10^{-3}$& 8.46 $\times 10^{-4}$ \\
& \textit{Coverage Greedy} &  & 4.28 & 0.479 & 5.93 $\times 10^{-3}$ & 1.12 $\times 10^{-3}$ \\
\end{tabular}
}\end{table}

Examining RMSE values in \Cref{fig:rmse_results}, we note that \textit{Greedy} improves relative to the other algorithms as prior uncertainty increases. 
We also see it outperform the heuristic greedy algorithms as $\Budget$ increases. 
In addition, we observe that \textit{Measurement Greedy} and \textit{Coverage Greedy} perform well for small cutoff values, which is explained by their explicit consideration of coverage. 
As the cutoff parameter $C$ increases, \textit{Greedy} and \textit{Random} perform similarly. 
Overall, \textit{Greedy} demonstrates strong performance across various different experimental setups, suggesting that its D-optimal cost function is a reliable proxy for localization accuracy. 
Finally, while \Cref{tab:algorithm_performance} shows that \textit{Greedy} takes considerably longer to run than its simpler competitors, it is still tractable for the offline design of sensing algorithms and has the potential to run much faster if implemented with optimized code or hardware acceleration.
\newline

\section{Conclusion} \label{sec:conclusion}
We have presented a novel formulation and solution to the problem of optimally placing range beacons in an autonomous mobile robot's workspace. 
Our approach leverages D-optimal Bayesian experimental design to reveal a submodular set function maximization problem with suboptimality guarantees for a fast greedy solver.
Preliminary results show that this method is able to improve the accuracy of localization over simple baseline methods in many simulated scenarios. 

In addition to conducting experiments with real data, future work will examine multiple measurement noise models which include line-of-sight modelling, try more complex position priors than the uniform isotropic distributions considered in this work, and consider full pose estimates as in \cite{goudar2023continuoustime}.
Additionally, if we remove the requirement that each position's prior is independently distributed or introduce a relative motion model representing noisy odometry measurements between consecutive positions, we can realistically model a larger variety of localization problems.
However, this change will remove the block diagonal structure of the Fisher information matrix exploited in the efficient evaluation of \Cref{eq:log_det}.
To make our approach tractable, we can investigate the convex relaxations used with E-optimal Bayesian experimental design in \cite{kaveti2023oasis} and \cite{doherty2022Spectral}.


\bibliographystyle{IEEEtran}
\bibliography{optimal-beacon-placement.bib}

\end{document}

%% file: packages.tex
\usepackage{times}

\usepackage{multicol}

\usepackage{xparse}
\usepackage{amsmath}
\usepackage{amssymb}
\usepackage{amsfonts}
\usepackage{dsfont}
\usepackage{graphicx}
\usepackage{url}
\usepackage{booktabs}
\usepackage[para]{threeparttable}
\usepackage{multirow}
\usepackage{makecell}
\usepackage{placeins}

\usepackage{amsthm}
\usepackage[colorlinks,bookmarksopen,bookmarksnumbered,citecolor=red,urlcolor=red]{hyperref}
\usepackage[capitalize]{cleveref}
\usepackage{xcolor}
\usepackage{framed}
\usepackage{caption}
\usepackage{subcaption}
\usepackage{pgfplots}
\usepackage{adjustbox}
\usepgfplotslibrary{groupplots,dateplot}
\usetikzlibrary{patterns,shapes.arrows}
\pgfplotsset{compat=newest}
\usepackage{flushend}

\usepackage{algorithm}
\usepackage{algpseudocode}
\makeatletter
\let\OldStatex\Statex
\renewcommand{\Statex}[1][0]{%
  \setlength\@tempdima{\algorithmicindent}%
  \OldStatex\hskip\dimexpr#1\@tempdima\relax}
\makeatother

\algnewcommand\algorithmicinput{\textbf{Input:}}
\algnewcommand\Input{\item[\algorithmicinput]}
\algnewcommand\algorithmicoutput{\textbf{Output:}}
\algnewcommand\Output{\item[\algorithmicoutput]}

\usepackage{setspace}
\usepackage[normalem]{ulem}

\colorlet{shadecolor}{gray!10}

\newtheorem{problem}{Problem}

\newtheorem{theorem}{Theorem}


%% file: notation.tex
\NewDocumentCommand\bbm{}{ \begin{bmatrix} }
\NewDocumentCommand\ebm{}{ \end{bmatrix} }
\NewDocumentCommand\Vector{m}{ \boldsymbol{\mathbf{#1}} }

\NewDocumentCommand\Matrix{m}{ \boldsymbol{\mathbf{#1}} }

\NewDocumentCommand\Real{}{ \mathbb{R} }
\NewDocumentCommand\Cardinality{m}{ \left\vert#1\right\vert }

\NewDocumentCommand\PSD{m}{ \mathbb{S}^{#1}_+ }
\NewDocumentCommand\PD{m}{ \mathbb{S}^{#1}_{++} }

\NewDocumentCommand\Indices{m}{[#1]}

\NewDocumentCommand\Expectation{mm}{ \mathbb{E}_{#2}\left[#1\right] }
\NewDocumentCommand\NormalDistribution{mm}{ \mathcal{N}\left(#1,#2\right) }

\NewDocumentCommand\Covariance{}{ \Matrix{\Sigma} }
\NewDocumentCommand\FIM{}{ \Matrix{\mathcal{I}} }

\NewDocumentCommand\Zero{}{ \Matrix{0} }

\NewDocumentCommand\Identity{}{ \Matrix{I} }

\NewDocumentCommand\Dim{}{ d }

\NewDocumentCommand\Estimate{m}{\hat{#1}}
\NewDocumentCommand\Prior{m}{\check{#1}}

\NewDocumentCommand\Defined{}{\triangleq}
\NewDocumentCommand\Define{}{ \triangleq }

\NewDocumentCommand\ArgMax{m}{ \operatorname*{argmax}_{#1} }

\NewDocumentCommand\Graph{}{ \mathcal{G} }
\NewDocumentCommand\Vertices{}{ \mathcal{V} }
\NewDocumentCommand\Edges{}{ \mathcal{E} }

\NewDocumentCommand\Neighbourhood{}{ \nu }

\newlength{\minuslength}
\NewDocumentCommand\NumBeacons{}{ N }
\NewDocumentCommand\NumPositions{}{ M }
\NewDocumentCommand\Budget{}{ K }
\NewDocumentCommand\FreeSpace{}{ \Omega }
\NewDocumentCommand\BeaconSet{}{ \mathcal{S} }
\NewDocumentCommand\CostFunction{}{ f_{\mathrm{D}} }
\NewDocumentCommand\CostFunctionNMS{}{ f_{\mathrm{D}}^+ }